\documentclass[journal]{IEEEtran}

\usepackage{amssymb}
\usepackage{graphicx}
\newcommand{\smallparagraph}[1]{ \vspace{3pt} \noindent \textbf{#1} }
\usepackage{bbm}

\usepackage{cite}

\usepackage[cmex10]{amsmath}
\DeclareMathOperator*{\argmax}{argmax}

\usepackage{algorithmicx}
\usepackage{algpseudocode}

\usepackage{url}

\usepackage{color}

\begin{document}

\title{Monte Carlo Search Algorithm Discovery\\ for One Player Games}

\author{Francis~Maes,~David~Lupien St-Pierre~and~Damien Ernst%
\thanks{Francis~Maes,~David~Lupien St-Pierre~and~Damien Ernst are at the Department of Electrical Engineering and Computer Science, University of Li\`ege, Li\`ege, Belgium.}
\thanks{E-mail: \{francis.maes, dlspierre, dernst\}@ulg.ac.be}

}

\markboth{IEEE Transactions on Computational Intelligence and AI in Games}%
{Monte-Carlo Search Algorithm Discovery for One Player Games}

\maketitle

\begin{abstract}
Much current research in AI and games is being devoted to Monte Carlo search (MCS) algorithms.
While the quest for a single unified MCS algorithm that would perform well on all problems is of major interest for AI, practitioners often know in advance the problem they want to solve, and spend plenty of time exploiting this knowledge to customize their MCS algorithm in a problem-driven way. We propose an MCS algorithm discovery scheme to perform this in an automatic and reproducible way. We first introduce a grammar over MCS algorithms that enables inducing a rich space of candidate algorithms. Afterwards, we search in this space for the algorithm that performs best on average for a given distribution of training problems. We rely on multi-armed bandits to approximately solve this optimization problem. The experiments, generated on three different domains, show that our approach enables discovering algorithms that outperform several well-known MCS algorithms such as Upper Confidence bounds applied to Trees and Nested Monte Carlo search. We also show that the discovered algorithms are generally quite robust with respect to changes in the distribution over the training problems.

\end{abstract}


\IEEEpeerreviewmaketitle

\section{Introduction}
\IEEEPARstart{M}{onte Carlo} search (MCS) algorithms rely on random simulations to evaluate the quality of states or actions in sequential decision making problems. Most of the recent progress in MCS algorithms has been obtained by integrating smart procedures to select the simulations to be performed. This has led to, among other things, the Upper Confidence bounds applied to Trees algorithm (UCT, \cite{kocsis2006bandit}) that was popularized thanks to breakthrough results in computer Go \cite{coulom2007efficient}.  This algorithm relies on a game tree to store simulation statistics and uses this tree to bias the selection of future simulations. While UCT is one way to combine random simulations with tree search techniques, many other approaches are possible. For example, the Nested Monte Carlo (NMC) search algorithm \cite{nestedMC2009caze}, which obtained excellent results in the last General Game Playing competition\footnote{\url{http://games.stanford.edu}} \cite{Mehat2010GGP}, relies on nested levels of search and does not require storing a game tree. 

How to best bias the choice of simulations is still an active topic in MCS-related research. Both UCT and NMC are attempts to provide generic techniques that perform well on a wide range of problems and that work with little or no prior knowledge. While working on such generic algorithms is definitely relevant to AI, MCS algorithms are in practice widely used in a totally different scenario, in which a significant amount of prior knowledge is available about the game or the sequential decision making problem to be solved.

People applying MCS techniques typically spend plenty of time exploiting their knowledge of the target problem so as to design more efficient problem-tailored variants of MCS. Among the many ways to do this, one common practice is automatic hyper-parameter tuning. By way of example, the parameter $C >0$ of UCT is in nearly all applications tuned through a more or less automated trial and error procedure. While hyper-parameter tuning is a simple form of problem-driven algorithm selection, most of the advanced algorithm selection work is done by humans, i.e., by researchers that modify or invent new algorithms to take the specificities of their problem into account.

The comparison and development of new MCS algorithms given a target problem is mostly a manual search process that takes much human time and is error prone. Thanks to modern computing power, automatic discovery is becoming a credible approach for partly automating this process. In order to investigate this research direction, we focus on the simplest case of (fully-observable) deterministic single-player games. Our contribution is twofold. First, we introduce a grammar over algorithms that enables generating a rich space of MCS algorithms. It also describes several well-known MCS algorithms, using a particularly compact and elegant description. 
Second, we propose a methodology based on multi-armed bandits for identifying the best MCS algorithm in this space, for a given distribution over training problems. We test our approach on three different domains. The results show that it often enables discovering new variants of MCS that significantly outperform generic algorithms such as UCT or NMC. We further show the good robustness properties of the discovered algorithms by slightly changing the characteristics of the problem.

This paper is structured as follows. Section \ref{sec:problem} formalizes the class of sequential decision making problems considered in this paper and formalizes the corresponding MCS algorithm discovery problem. Section \ref{sec:grammar} describes our grammar over MCS algorithms and describes several well-known MCS algorithms in terms of this grammar. Section \ref{sec:bandit} formalizes the search for a good MCS algorithm as a multi-armed bandit problem. We experimentally evaluate our approach on different domains in Section \ref{sec:experiments}. Finally, we discuss related work in Section \ref{sec:related} and conclude in Section \ref{sec:conclusion}.

\section{Problem statement}
\label{sec:problem}
\newcommand{\expect}[1]{\textsf{E}\{#1\}}
\newcommand{\expectwrt}[2]{\textsf{E}_{#2}\{#1\}}

We consider the class of finite-horizon fully-observable deterministic sequential decision-making problems. A problem $P$ is a triple $(x_1, f, g)$ where $x_1 \in \mathcal{X}$ is the initial state, $f$ is the transition function, and $g$ is the reward function. The dynamics of a problem is described by
\begin{eqnarray}
x_{t+1}=f(x_t,u_t)  \quad  t=1,2,\ldots,T,
\end{eqnarray}
\noindent where for all $t$, the state $x_t$ is an element of the state space $\mathcal{X}$ and the action $u_t$ is an element of the action space. We denote by $\mathcal{U}$ the whole action space and by $\mathcal{U}_x \subset \mathcal{U}$ the subset of actions which are available in state $x \in \mathcal{X}$. In the context of one player games, $x_t$ denotes the current state of the game and $\mathcal{U}_{x_t}$ are the legal moves in that state. We make no assumptions on the nature of $\mathcal{X}$ but assume that $\mathcal{U}$ is finite. We assume that when starting from $x_1$, the system enters a final state after $T$ steps and we denote by $\mathcal{F} \subset \mathcal{X}$ the set of these final states\footnote{In many problems, the time at which the game enters a final state is not fixed, but depends on the actions played so far. It should however be noted that it is possible to make these problems fit this fixed finite time  formalism by postponing artificially the end of the game until $T$. This can be done, for example, by considering that when the game ends before $T$, a ``pseudo final state'' is reached from which, whatever the actions taken, the game will reach the real final state in $T$.}. Final states $x \in \mathcal{F}$ are associated to rewards $g(x) \in \mathbb R$ that should be maximized.

A search algorithm $A(\cdot)$ is a stochastic algorithm that explores the possible sequences of actions to approximately maximize
\begin{eqnarray}
   A(P = (x_1, f, g) ) \simeq \argmax_{u_1, \dots, u_T} g(x_{T+1}) \ ,
\end{eqnarray}
subject to $x_{t+1}=f(x_t, u_t)$ and $u_t \in \mathcal{U}_{x_t}$. In order to fulfill this task, the algorithm is given a finite amount of computational time, referred to as the \textit{budget}. To facilitate reproducibility, we focus primarily in this paper on a budget expressed as the maximum number $B > 0$ of sequences $(u_1, \dots, u_T)$ that can be evaluated, or, equivalently, as the number of calls to the reward function $g(\cdot)$. Note, however, that it is trivial in our approach to replace this definition by other budget measures, as illustrated in one of our experiments in which the budget is expressed as an amount of CPU time.

We express our prior knowledge  as a distribution over problems $\mathcal{D}_P$, from which we can sample any number of \textit{training problems} $P \sim \mathcal{D}_P$. The quality of a search algorithm $A^B(\cdot)$ with budget $B$ on this distribution is denoted by $J^B_A(\mathcal{D}_P)$ and is defined as the expected quality of solutions found on problems drawn from $\mathcal{D}_P$:
\begin{eqnarray}
  J^B_A(\mathcal{D}_P) = \expectwrt{ \expectwrt{g(x_{T+1})}{x_{T+1} \sim A^B(P)} }{P \sim \mathcal{D}_P} \ ,
\end{eqnarray}
where $x_{T+1} \sim A^B(P)$ denotes the final states returned by algorithm $A$ with budget $B$ on problem $P$.

Given a class of candidate algorithms $\mathcal{A}$ and given the budget $B$, the algorithm discovery problem amounts to selecting an algorithm $A^* \in \mathcal{A}$ of maximal quality:
\begin{eqnarray}
\label{eq:objective}
  A^* = \argmax_{A \in \mathcal{A}} J^B_A(\mathcal{D}_P) \ .
\end{eqnarray}

The two main contributions of this paper are: (i) a grammar that enables inducing a rich space $\mathcal{A}$  of candidate MCS algorithms, and (ii) an efficient procedure to approximately solve Eq.~\ref{eq:objective}.

\section{A grammar for Monte-Carlo search algorithms}
\label{sec:grammar}
\newcommand{\tcurrent}{\ensuremath{t}}
\newcommand{\tloop}{\ensuremath{\tau}}

All MCS algorithms share some common underlying general principles: random simulations, look-ahead search, time-receding control, and bandit-based selection. The grammar that we introduce in this section aims at capturing these principles in a pure and atomic way. We first give an overall view of our approach, then present in detail the components of our grammar, and finally describe previously proposed algorithms by using this grammar.

\subsection{Overall view}

We call \textit{search components} the elements on which our grammar operates. Formally, a search component is a stochastic algorithm that, when given a partial sequence of actions $(u_1, \dots, u_{\tcurrent-1})$, generates one or multiple completions $(u_\tcurrent, \dots, u_T)$ and evaluates them using the reward function $g(\cdot)$. The search components are denoted by $S \in \mathcal{S}$, where $\mathcal{S}$ is the space of all possible search components.

Let $S$ be a particular search component. We define the search algorithm $A_S \in \mathcal{A}$ as the algorithm that, given the problem $P$, executes $S$ repeatedly with an empty partial sequence of actions $()$, until the computational budget is exhausted. The search algorithm $A_S$ then returns the sequence of actions $(u_1, \dots, u_T)$ that led to the highest reward $g(\cdot)$. 

In order to generate a rich class of search components---hence a rich class of search algorithms---in an inductive way, we rely on search-component \textit{generators}. Such generators are functions $\Psi : \Theta \rightarrow \mathcal{S}$ that define a search component $S=\Psi(\theta) \in \mathcal{S}$ when given a set of parameters $\theta \in \Theta$. Our grammar is composed of five search component generators that are defined in Section \ref{ssec:components}: $\Psi \in \{simulate, repeat, lookahead, step, select\}$. 
Four of these search component generators are parametrized by sub-search components. For example, $step$ and $lookahead$ are functions $\mathcal{S} \rightarrow \mathcal{S}$. These functions can be nested recursively to generate more and more evolved search components. We construct the space of search algorithms $\mathcal{A}$ by performing this in a systematic way, as detailed in Section \ref{ssec:algospace}.

\subsection{Search components}
\label{ssec:components}

Figure~\ref{alg:searchprocs} describes our five search component generators. Note that we distinguish between search component \textit{inputs} and search component generator \textit{parameters}. All our search components have the same two inputs: the sequence of already decided actions $(u_1, \dots, u_{\tcurrent-1})$ and the current state $x_\tcurrent \in \mathcal{X}$. The parameters differ from one search component generator to another. For example, $simulate$ is parametrized by a simulation policy $\pi^{simu}$ and $repeat$ is parametrized by the number of repetitions $N > 0$ and by a sub-search component. We now give a detailed description of these search component generators.

\begin{figure}[h!]
\caption{Search component generators}
\begin{algorithmic}

\medskip
\State{\large{\textsc{Simulate}}($(u_1, \dots, u_{\tcurrent-1}), x_\tcurrent$)}
\State{\textbf{Param:} $\pi^{simu} \in \Pi^{simu}$}
\smallskip
\For{$\tloop=\tcurrent \mbox{ to } T$}
  \State{$u_\tloop \sim \pi^{simu}(x_\tloop)$} 
  \State{$x_{\tloop+1} \leftarrow f(x_\tloop, u_\tloop)$}
\EndFor
\State{\textsc{yield}($(u_1, \dots, u_T)$)}

\medskip
\State{------------------------}
\medskip

\State{\large{\textsc{Repeat}}($(u_1, \dots, u_{\tcurrent-1}), x_\tcurrent$)}
\State{\textbf{Param:} $N > 0, S \in \mathcal{S}$}
\smallskip

\For{$i=1 \mbox{ to } N$}
  \State{\textsc{invoke}($S, (u_1, \dots, u_{\tcurrent-1}), x_\tcurrent$)}
\EndFor

\medskip
\State{------------------------}
\medskip

\State{\large{\textsc{LookAhead}}($(u_1, \dots, u_{\tcurrent-1}), x_\tcurrent$)}
\State{\textbf{Param:} $S \in \mathcal{S}$}
\smallskip
\For{$u_{\tcurrent} \in \mathcal{U}_{x_\tcurrent}$}
  \State{$x_{\tcurrent+1} \leftarrow f(x_\tcurrent, u_\tcurrent)$}
  \State{\textsc{invoke}($S, (u_1, \dots, u_\tcurrent), x_{\tcurrent+1}$)}
\EndFor

\medskip
\State{------------------------}
\medskip

\State{\large{\textsc{Step}}($(u_1, \dots, u_{\tcurrent-1}), x_\tcurrent$)}
\State{\textbf{Param:} $S \in \mathcal{S}$}
\smallskip

\For{$\tloop=\tcurrent \mbox{ to } T$}
  \State{\textsc{invoke}($S, (u_1, \dots, u_{\tloop-1}), x_\tloop$)}
  \State{$u_\tloop \leftarrow u^*_\tloop$}
  \State{$x_{\tloop+1} \leftarrow f(x_\tloop, u_\tloop)$}
\EndFor

\medskip
\State{------------------------}
\medskip

\State{\large{\textsc{Select}}($(u_1, \dots, u_{\tcurrent-1}), x_\tcurrent$)}
\State{\textbf{Param:} $\pi^{sel} \in \Pi^{sel}, S \in \mathcal{S}$}
\smallskip
  
  \For{$\tloop=\tcurrent \mbox{ to } T$} \Comment{Select}
    \State $u_\tloop \sim \pi^{sel}(x)$
    \State $x_{\tloop+1} \leftarrow f(x_\tloop, u_\tloop)$
    \If{$n(x_{\tloop+1}) = 0$}
      \State{\textbf{break}}
    \EndIf
  \EndFor
  \State{$t_{leaf} \leftarrow \tloop$}

  \medskip  
  
  \State{\textsc{invoke}($S, (u_1, \dots, u_{t_{leaf}}), x_{t_{leaf}+1}$)} \Comment{Sub-search}

  \medskip
      
  \For{$\tloop=t_{leaf} \mbox{ to } 1$}   \Comment{Backpropagate}
    \State{$n(x_{\tloop+1}) \leftarrow n(x_{\tloop+1}) + 1$}  
    \State{$n(x_\tloop, u_\tloop) \leftarrow n(x_\tloop, u_\tloop) + 1$}
    \State{$s(x_\tloop, u_\tloop) \leftarrow s(x_\tloop, u_\tloop) + r^*$}
  \EndFor
  \State{$n(x_1) \leftarrow n(x_1) + 1$}
\end{algorithmic}
\label{alg:searchprocs}
\end{figure}

\begin{figure}[tb]
\caption{Yield and invoke commands}
\begin{algorithmic}
\Require{$g : \mathcal{F} \rightarrow \mathbb R$, the reward function}
\Require{$B > 0$, the computational budget}
\State{Initialize global: $numCalls \leftarrow 0$}
\State{Initialize local: $r^* \leftarrow -\infty$}
\State{Initialize local: $(u^*_1, \dots, u^*_T) \leftarrow \emptyset$}
\Procedure{Yield}{$(u_1, \dots, u_T)$}
  \State{$r = g(x)$}
  \If{$r > r^*$}
     \State{$r^* \leftarrow r$}
     \State{$(u^*_1, \dots, u^*_T) \leftarrow (u_1, \dots, u_T)$}
  \EndIf
  \State{$numCalls \leftarrow numCalls + 1$}
   \If{$numCalls = B$}
     \State{\textbf{stop search}}
   \EndIf
\EndProcedure

\smallskip

\Procedure{Invoke}{$S \in \mathcal{S}, (u_1, \dots, u_{\tcurrent-1}) \in \mathcal{U}^*, x_\tcurrent \in \mathcal{X}$}
\If{$\tcurrent \leq T$}
  \State{$S((u_1, \dots, u_{\tcurrent-1}), x_\tcurrent)$}
\Else
  \State{\textbf{yield} $(u_1, \dots, u_T)$}
\EndIf
\EndProcedure

\end{algorithmic}
\label{alg:invoke}
\end{figure}

\smallparagraph{Simulate} The $simulate$ generator is parametrized by a policy $\pi^{simu} \in \Pi^{simu}$ which is a stochastic mapping from states to actions: $u \sim \pi^{simu}(x)$. In order to generate the completion $(u_\tcurrent, \dots, u_T)$, $simulate(\pi^{simu})$ repeatedly samples actions $u_\tloop$ according to $\pi^{simu}(x_\tloop)$ and performs transitions $x_{\tloop + 1} = f(x_\tloop, u_\tloop)$ until reaching a final state.
A default choice for the simulation policy is the uniformly random policy, defined as
\begin{eqnarray}
  \expect{\pi^{random}(x) = u} = \begin{cases}
        \frac{1}{|\mathcal{U}_x|} & \mbox{if } u \in \mathcal{U}_x \\
        0 & \mbox{otherwise.} \\
     \end{cases}
\end{eqnarray}
Once the completion $(u_\tcurrent, \dots, u_T)$ is fulfilled, the whole sequence $(u_1, \dots, u_T)$ is \textit{yielded}. This operation is detailed in Figure~\ref{alg:invoke} and proceeds as follows: (i) it computes the reward of the final state $x_{T+1}$, (ii) if the reward is larger than the largest reward found previously, it replaces the best current solution, and (iii) if the budget $B$ is exhausted, it stops the search. 

Since algorithm $A_P$ repeats $P$ until the budget is exhausted, the search algorithm $A_{simulate(\pi^{simu})} \in \mathcal{A}$ is the algorithm that samples $B$ random trajectories $(u_1, \dots, u_T)$, evaluates each of the final state rewards $g(x_{T+1})$, and returns the best found final state. This simple random search algorithm is sometimes called \textit{Iterative Sampling} \cite{Tesauro96MCS}.

Note that, in the \textsc{yield} procedure, the variables relative to the best current solution ($r^*$ and $(u^*_1, \dots, u^*_T)$) are defined locally for each search component, whereas the $numCalls$ counter is global to the search algorithm. This means that if $S$ is a search component composed of different nested levels of search (see the examples below), the best current solution is kept in memory at each level of search.

\smallparagraph{Repeat} Given a positive integer $N > 0$ and a search component $S \in \mathcal{S}$, $repeat(N, S)$ is the search component that repeats $N$ times the search component $S$. For example, $S = repeat(10, simulate(\pi^{simu}))$ is the search component that draws $10$ random simulations using $\pi^{simu}$. The corresponding search algorithm $A_S$ is again iterative sampling, since search algorithms repeat their search component until the budget is exhausted. In Figure~\ref{alg:searchprocs}, we use the \textsc{invoke} operation each time a search component calls a sub-search component. This operation is detailed in Figure~\ref{alg:invoke} and ensures that no sub-search algorithm is called when a final state is reached, i.e., when $\tcurrent=T+1$.

\smallparagraph{Look-ahead} For each legal move $u_\tcurrent \in \mathcal{U}_{x_\tcurrent}$, $lookahead(S)$ computes the successor state $x_{\tcurrent+1} = f(x_\tcurrent, u_\tcurrent)$ and runs the sub-search component $S \in \mathcal{S}$ starting from the sequence $(u_1, \dots, u_\tcurrent)$. For example, $lookahead(simulate(\pi^{simu}))$ is the search component that, given the partial sequence $(u_1, \dots, u_{t-1})$, generates one random trajectory for each legal next action $u_t \in \mathcal{U}_{x_t}$. Multiple-step look-ahead search strategies naturally write themselves with nested calls to $lookahead$. As an example, $lookahead(lookahead(lookahead(simulate(\pi^{simu}))))$ is a search component that runs one random trajectory per legal combination of the three next actions $(u_t, u_{t+1}, u_{t+2})$.

\smallparagraph{Step} For each remaining time step $\tloop \in [\tcurrent, T]$, $step(S)$ runs the sub-search component $S$, extracts the action $u_\tloop$ from $(u^*_1, \dots, u^*_T)$ (the best currently found action sequence, see Figure~\ref{alg:invoke}), and performs transition $x_{\tloop+1} = f(x_\tloop, u_\tloop)$. The search component generator $step$ enables implementing time receding search mechanisms, e.g., $step(repeat(100, simulate(\pi^{simu})))$ is the search component that selects the actions $(u_1, \dots, u_T)$ one by one, using 100 random trajectories to select each action. As a more evolved example, \begin{small}$step(lookahead(lookahead(repeat(10, simulation(\pi^{simu})))))$\end{small} is a time receding strategy that performs $10$ random simulations for each two first actions $(u_t,u_{t+1})$ to decide which action $u_t$ to select.

\smallparagraph{Select} This search component generator implements most of the behaviour of a Monte Carlo Tree Search (MCTS, \cite{kocsis2006bandit}). It relies on a game tree, which is a non-uniform look-ahead tree with nodes corresponding to states and edges corresponding to transitions. The role of this tree is twofold: it stores statistics on the outcomes of sub-searches  and it is used to bias sub-searches towards promising sequences of actions. A search component $select(\pi^{sel}, S)$ proceeds in three steps: the \textit{selection} step relies on the statistics stored in the game tree to select a (typically small) sub-sequence of actions $(u_{\tcurrent}, \dots, u_{t_{leaf}})$, the \textit{sub-search} step invokes the sub-search component $S \in \mathcal{S}$ starting from $(u_1, \dots, u_{t_{leaf}})$, and the \textit{backpropagation} step updates the statistics to take into account the sub-search result. 

We use the following notation to denote the information stored by the look-ahead tree:  $n(x, u)$ is the number of times the  action $u$ was selected in state $x$, $s(x, u)$ is the sum of rewards that were obtained when running \textit{sub-search} after having selected action $u$ in state $x$, and $n(x)$ is the number of times state $x$ was selected: $n(x) = \sum_{u\in\mathcal{U}_x} n(x,u)$. In order to quantify the quality of a sub-search, we rely on the reward of the best solution that was tried during that sub-search:  $r^* = \max g(x)$. In the simplest case, when the sub-search component is $S = simulate(\pi^{simu})$, $r^*$ is the reward associated to the final state obtained by making the random simulation with policy $\pi^{simu}$, as usual in MCTS. In order to select the first actions, \textit{selection} relies on a \textit{selection policy} $\pi^{sel} \in \Pi^{sel}$, which is a stochastic function that, when given all stored information related to state $x$ (i.e., $n(x)$, $n(x, u)$, and $s(x, u), \forall u \in \mathcal{U}_x$), selects an action $u \in \mathcal{U}_x$. The selection policy has two contradictory goals to pursue: \textit{exploration}, trying new sequences of actions to increase knowledge, and \textit{exploitation}, using current knowledge to bias computational efforts towards promising sequences of actions.
Such exploration/exploitation dilemmas are usually formalized as a multi-armed bandit problem, hence $\pi^{sel}$ is typically one of policies commonly found in the multi-armed bandit literature. The probably most well-known such policy is UCB-1 \cite{Auer02bandits}:
\begin{eqnarray}
\label{eq:ucb}
  \pi^{ucb-1}_C(x) = \argmax_{u \in \mathcal{U}_x} \frac{s(x,u)}{n(x,u)} + C \sqrt{\frac{\ln n(x)}{n(x,u)}} \ ,
\end{eqnarray}
where division by zero returns $+\infty$ and where $C > 0$ is a hyper-parameter that enables the control of the exploration / exploitation tradeoff. 

\subsection{Description of previously proposed algorithms}
\label{sec:genericAlgo}

Our grammar enables generating a large class of MCS algorithms, which includes several already proposed algorithms. We now overview these algorithms, which can be described particularly compactly and elegantly thanks to our grammar:
\begin{itemize}
  \item The simplest Monte Carlo algorithm in our class is \textit{Iterative Sampling}. This algorithm draws random simulations until the computational time is elapsed and returns the best solution found:
\begin{eqnarray}
  is = simulate(\pi^{simu}).
\end{eqnarray}
  \item In general, iterative sampling  is used during a certain time to decide which action to select (or which move to play) at each step of the decision problem. The corresponding search component is
  \begin{eqnarray}
  is^\prime = step(repeat(N, simulate(\pi^{simu}))),
\end{eqnarray}
where $N$ is the number of simulations performed for each decision step.
  \item The \textit{Reflexive Monte Carlo} search algorithm introduced in \cite{cazenave2007reflexive} proposes using a Monte Carlo search of a given level to improve the search of the upper level. The proposed algorithm can be described as follows:
 \begin{multline}
   rmc(N_1, N_2) = step(repeat(N_1, \\ step(repeat(N_2,  simulate(\pi^{simu}))))),
 \end{multline}
 where $N_1$ and $N_2$ are called the number of meta-games and the number of games, respectively.
 \item The Nested Monte Carlo (NMC) search algorithm \cite{nestedMC2009caze} is a recursively defined algorithm generalizing the ideas of Reflexive Monte Carlo search. NMC can be described in a very natural way by our grammar. The basic search level $l=0$ of NMC simply performs a random simulation:
\begin{eqnarray}
  nmc(0) = simulate(\pi^{random}) \ .
\end{eqnarray}
The level $l > 0$ of NMC relies on level $l-1$ in the following way:
\begin{eqnarray}
  nmc(l) = step(lookahead(nmc(l - 1))) \ .
\end{eqnarray}

\item Single-player MCTS \cite{Chaslot2006MC,Schadd08single-playermonte-carlo, demesmay2009ICML} selects actions one after the other. In order to select one action, it relies on \textit{select} combined with random simulations. The corresponding search component is thus
\begin{multline}
  mcts(\pi^{sel},\pi^{simu}, N) = step( repeat(N, \\ select(\pi^{sel}, simulate(\pi^{simu})))) \ ,
\end{multline}
where $N$ is the number of iterations allocated to each decision step. UCT is one of the best known variants of MCTS. It relies on the $\pi^{ucb-1}_C$ selection policy and is generally used with a uniformly random simulation policy:
\begin{eqnarray}
  uct(C, N) = mcts(\pi^{ucb-1}_C, \pi^{random}, N) \ .
\end{eqnarray}

\item In the spirit of the work on nested Monte Carlo, the authors of  \cite{Chaslot2009MetaMC} proposed the Meta MCTS approach, which replaces the simulation part of an upper-level MCTS algorithm by a whole lower-level MCTS algorithm. While they presented this approach in the context of two-player games, we can describe its equivalent for one-player games with our grammar:
\begin{multline}
  metamcts(\pi^{sel}, \pi^{simu}, N_1, N_2) = \\
  	step( repeat(N_1, select(\pi^{sel}, \\ mcts(\pi^{sel}, \pi^{simu}, N_2)) \,
\end{multline}
where $N_1$ and $N_2$ are the budgets for the higher-level and lower-level MCTS algorithms, respectively.
\end{itemize}

In addition to offering a framework for describing these already proposed algorithms, our grammar enables generating a large number of new hybrid MCS variants. We give, in the next section, a procedure to automatically identify the best such variant for a given problem.

\section{Bandit-based algorithm discovery}
\label{sec:bandit}
We now move to the problem of solving Eq.~\ref{eq:objective}, i.e., of finding, for a given problem, the best algorithm $A$ from among a large class $\mathcal{A}$ of algorithms derived with the grammar previously defined. Solving this algorithm discovery problem exactly is impossible in the general case since the objective function involves two infinite expectations: one over the problems $P \sim \mathcal{D}_P$ and another over the outcomes of the algorithm. In order to approximately solve Eq.~\ref{eq:objective}, we adopt the formalism of multi-armed bandits and proceed in two steps: we first construct a finite set of candidate algorithms $\mathcal{A}_{D,\Gamma} \subset \mathcal{A}$ (Section \ref{ssec:algospace}), and then treat each of these algorithms as an arm and use a multi-armed bandit policy to select how to allocate computational time to the performance estimation of the different algorithms (Section \ref{ssec:algomab}). It is worth mentioning that this two-step approach follows a general methodology for automatic discovery that we already successfully applied to multi-armed bandit policy discovery \cite{Maes2012Icaart,Maes2012IcaartLong}, reinforcement learning policy discovery \cite{Castronovo2012Ewrl}, and optimal control policy discovery \cite{Maes2012PolicySearch}.

\subsection{Construction of the algorithm space}
\label{ssec:algospace}

We measure the complexity of a search component $S \in \mathcal{S}$ using its \textit{depth}, defined as the number of nested search components constituting $S$, and denote this quantity by $depth(S)$. For example, $depth(simulate(\pi^{simu}))$ is 1, $depth(uct)$ is $4$, and $depth(nmc(3))$ is $7$. 

Note that $simulate$, $repeat$, and $select$ have parameters which are not search components: the simulation policy $\pi^{simu}$, the number of repetitions $N$, and the selection policy $\pi^{sel}$, respectively. In order to generate a finite set of algorithms using our grammar, we rely on predefined finite sets of possible values for each of these parameters. We denote by $\Gamma$ the set of these finite domains. The discrete set $\mathcal{A}_{D,\Gamma}$ is constructed by enumerating all possible algorithms up to depth $D$ with constants $\Gamma$, and is pruned using the following rules:
\begin{itemize}
  \item \textit{Canonization of repeat:} Both search components $S_1 = step(repeat(2, repeat(5, S_{sub})))$ and $S_2 = step(repeat(5, repeat(2, S_{sub})))$ involve running $S_{sub}$ $10$ times at each step. In order to avoid having this kind of algorithm duplicated, we collapse nested $repeat$ components into single $repeat$ components. With this rule, $S_1$ and $S_2$ both reduce to $step(repeat(10, S_{sub}))$.
  \item \textit{Removal of nested selects:} A search component such as $select(\pi^{sel}, select(\pi^{sel}, S))$ is ill-defined, since the inner $select$ will be called with a different initial state $x_t$ each time, making it  behave randomly. We therefore exclude search components involving two directly nested $select$s.
  \item \textit{Removal of repeat-as-root:} Remember that the MCS algorithm $A_S \in \mathcal{A}$ runs $S$ repeatedly until the computational budget is exhausted. Due to this repetition, algorithms such as $A_{simulate(\pi^{simu})}$ and $A_{repeat(10, simulate(\pi^{simu}))}$ are equivalent. To remove these duplicates, we reject all search components whose ``root'' is $repeat$.
\end{itemize}

\begin{table}
\begin{tabular}{c|cc}
Depth 1--2 & \multicolumn{2}{c}{Depth 3} \\
\hline
 sim & lookahead(repeat(2, sim)) & step(repeat(2, sim)) \\
 & lookahead(repeat(10, sim)) & step(repeat(10, sim)) \\
lookahead(sim) & lookahead(lookahead(sim)) & step(lookahead(sim)) \\
step(sim)  & lookahead(step(sim)) & step(step(sim))  \\
select(sim) & lookahead(select(sim))  & step(select(sim)) \\
 & select(repeat(2, sim)) & select(repeat(10, sim)) \\
 & select(lookahead(sim)) & select(step(sim)) \\
\hline
\end{tabular}
\caption{Unique algorithms up to depth 3}
\label{tbl:AlgorithmsUpToDepth3}
\end{table}

\newcommand{\bandits}{\ensuremath{\nu}}
 In the following, $\bandits$ denote the cardinality of the set of candidate algorithms: $\mathcal{A}_{D,\Gamma} = \left\{ A_1, \ldots, A_\bandits  \right\}$. To illustrate the construction of this set, consider a simple case where  the maximum depth is $D=3$ and where the constants $\Gamma$ are $\pi^{simu} = \pi^{random}, N \in \{2,10\}$, and $\pi^{sel} = \pi^{ucb-1}_C$.  The corresponding space  $\mathcal{A}_{D,\Gamma}$ contains $\bandits=18$ algorithms. These algorithms are  given in Table \ref{tbl:AlgorithmsUpToDepth3}, where we use $sim$ as an abbreviation for $simulate(\pi^{simu})$.

\subsection{Bandit-based algorithm discovery}
\label{ssec:algomab}

One simple approach to approximately solve Eq.~\ref{eq:objective} is to estimate the objective function through an empirical mean computed using a finite set of training problems $\{ P^{(1)}, \dots, P^{(M)} \}$, drawn from $\mathcal{D}_P$:
\begin{eqnarray}
  J^B_A(\mathcal{D}_P) \simeq \frac{1}{M} \sum_{i=1}^{M} g(x_{T+1}) | x_{T+1} \sim A^B(P^{(i)}) \ ,
\end{eqnarray}
where $x_{T+1}$ denotes one outcome of algorithm $A$ with budget $B$ on problem  $P^{(i)}$. To solve Eq.~\ref{eq:objective}, one can then compute this approximated objective function for all algorithms $A \in \mathcal{A}_{D,\Gamma}$ and simply return the algorithm with the highest score. While extremely simple to implement, such an approach often requires an excessively large number of samples $M$ to work well, since the variance of $g(\cdot)$ may be quite large.

In order to optimize Eq.~\ref{eq:objective} in a smarter way, we propose to formalize this problem as a multi-armed bandit problem. To each algorithm  $A_k \in \mathcal{A}_{D,\Gamma}$, we associate an arm. Pulling the arm $k$ for the $t_k$th time involves selecting the problem $P^{(t_k)}$ and running the algorithm $A_k$ once on this problem. This leads to a reward associated to arm $k$ whose value is the reward $g(x_{T+1})$ that comes with the solution $x_{T+1}$ found by algorithm $A_k$. The purpose of multi-armed bandit algorithms is to process the sequence of observed rewards to select in a smart way the next algorithm to be tried, so that when the time allocated to algorithm discovery is exhausted, one (or several) high-quality algorithm(s) can be identified.  How to select arms so as to identify the best one in a finite amount of time is known as the \textit{pure exploration} multi-armed bandit problem \cite{Bubeck09pure}.  
It has been shown that index based policies based on upper confidence bounds such as UCB-1 were also good policies for solving pure exploration bandit problems. Our optimization procedure works thus by repeatedly playing arms according to such a policy. In our experiments, we perform a fixed number of such iterations. In practice this multi-armed bandit approach can provide an answer at anytime, returning the algorithm $A_k$ with the currently highest empirical reward mean.

\subsection{Discussion}

Note that other approaches could be considered for solving our algorithm discovery problem. In particular, optimization over expression spaces induced by a grammar such as ours is often solved using Genetic Programming (GP) \cite{Koza2005GP}. GP works by evolving a population of solutions, which, in our case, would be MCS algorithms. At each iteration, the current population is evaluated, the less good solutions are removed, and the best solutions are used to construct new candidates using mutation and cross-over operations. Most existing GP algorithms assume that the objective function is (at least approximately) deterministic. One major advantage of the bandit-based approach is to natively take into account the stochasticity of the objective function and its decomposability into problems. Thanks to the bandit formulation, badly performing algorithms are quickly rejected and the computational power is more and more focused on the most promising algorithms. 

The main strengths of our bandit-based approach are the following. First, it is simple to implement and does not require entering into the details of complex mutation and cross-over operators. Second, it has only one hyper-parameter (the exploration/exploitation coefficient). Finally, since it is based on exhaustive search and on multi-armed bandit theory, formal guarantees can easily be derived to bound the regret, i.e., the difference between the performance of the best algorithm and the performance of the algorithm discovered \cite{Auer2002,Bubeck09pure,Coquelin2007BanditTS}. 

Our approach is restricted to relatively small depths $D$ since it relies on exhaustive search. In our case, we believe that many interesting MCS algorithms can be described using search components with low depth. In our experiments, we used $D=5$, which already provides many original hybrid algorithms that deserve further research. Note that GP algorithms do not suffer from such a limit, since they are able to generate deep and complex solutions through mutation and cross-over of smaller solutions. If the limit $D=5$ was too restrictive, a major way of improvement would thus consist in combining the idea of bandits with those of GP. In this spirit, the authors of \cite{Hoock2010BBGP}  recently proposed a hybrid approach in which the selection of the members of a new population is posed as a multi-armed bandit problem. This enables combining the best of the two approaches: multi-armed bandits enable taking natively into account the stochasticity and decomposability of the objective function, while GP cross-over and mutation operators are used to generate new candidates dynamically in a smart way.

\section{Experiments}
\label{sec:experiments}
We now apply our automatic algorithm discovery approach to three different testbeds: Sudoku, Symbolic Regression, and Morpion Solitaire. The aim of our experiments was to show that our approach discovers MCS algorithms that outperform several generic (problem independent) MCS algorithms: outperforms them  on the training instances, on new testing instances, and even on instances drawn from distributions different from the original distribution used for the learning.

We first describe the experimental protocol in Section \ref{protocol}. We perform a detailed study of the behavior of our approach applied to the Sudoku domain in Section \ref{sec:sudoku}. Section \ref{sec:symbreg}, and \ref{sec:morpion} then give the results obtained on the other two domains. Finally, Section \ref{sec:xp:dis} gives an overall discussion of our results.

\subsection{Protocol}
\label{protocol}

We now describe the experimental protocol that will be used in the remainder of this section.

\smallparagraph{Generic algorithms} The generic algorithms are Nested Monte Carlo, Upper Confidence bounds applied to Trees, Look-ahead Search, and Iterative sampling. The search components for Nested Monte Carlo ($nmc$), UCT ($uct$), and Iterative sampling ($is$) have already been defined in Section \ref{sec:genericAlgo}.  The search component for Look-ahead Search of level $l>0$ is defined by $la(l) = step(larec(l))$, where 
\begin{eqnarray}
  larec(l) = \begin{cases} 
      lookahead(larec(l-1)) & \mbox{if $l>0$} \\
      simulate(\pi^{random}) & \mbox{otherwise.} \\
   \end{cases}
\end{eqnarray}

For both $la(\cdot)$ and $nmc(\cdot)$, we try all values within the range $[1,5]$ for the level parameter. Note that $la(1)$ and $nmc(1)$ are equivalent, since both are defined by the search component $step(lookahead(simulate(\pi^{random})))$.  For $uct(\cdot)$, we try the following values of $C$: $\{ 0, 0.3, 0.5, 1.0 \}$ and set the budget per step to $\frac{B}{T}$, where $B$ is the total budget and $T$ is the horizon of the problem. This leads to the following set of generic algorithms: $\{ nmc(2)$, $nmc(3)$, $nmc(4)$, $nmc(5)$, $is$, $la(1)$, $la(2)$, $la(3)$, $la(4)$, $la(5)$, $uct(0)$, $uct(0.3)$, $uct(0.5)$, and $uct(1) \}$. Note that we omit the $\frac{B}{T}$ parameter in $uct$ for the sake of conciseness.

\smallparagraph{Discovered algorithms} In order to generate the set of candidate algorithms, we used the following constants $\Gamma$: $repeat$ can be used with $2,5,10$, or $100$ repetitions; and $select$ relies on the $UCB1$ selection policy from Eq.~\eqref{eq:ucb} with the constants $\{0, 0.3, 0.5, 1.0\}$. We create a pool of algorithms by exhaustively generating all possible combinations of the search components up to depth $D=5$. We apply the pruning rules described in Section \ref{ssec:algospace}, which results in a set of $\bandits = 3,155$ candidate MCS algorithms. 

\smallparagraph{Algorithm discovery} In order to carry out the algorithm discovery, we used a UCB policy for $100 \times \bandits$ time steps, i.e., each candidate algorithm was executed $100$ times on average. As discussed in Section \ref{ssec:algomab}, each bandit step involves running one of the candidate algorithms on a problem $P \sim \mathcal{D}_P$. We refer to $\mathcal{D}_P$ as the \textit{training distribution} in the following. Once we have played the UCB policy for $100 \times \bandits$ time steps, we sort the algorithms by their average training performance and report the ten best algorithms.

\smallparagraph{Evaluation} Since algorithm discovery is a form of ``learning from examples'', care must be taken with overfitting issues. Indeed, the discovered algorithms may perform well on the training problems $P$ while performing poorly on other problems drawn from  $\mathcal{D}_P$. Therefore, to evaluate the MCS algorithms, we used a set of $10,000$ \textit{testing problems} $P \sim \mathcal{D}_P$ which are different from the training problems. We then evaluate the score of an algorithm as the mean performance obtained when running it once on each testing problem.

In each domain, we futher test the algorithms either by changing the budget $B$ and/or by using a new distribution $\mathcal{D}^\prime_P$ that differs from the training distribution $\mathcal{D}_P$. In each such experiment, we draw $10,000$ problems from $\mathcal{D}^\prime_P$ and run the algorithm once on each problem. 

In one domain (Morpion Solitaire), we used a particular case of our general setting, in which there was a single training problem $P$, i.e., the distribution $\mathcal{D}_P$ was degenerate and always returned the same $P$. In this case, we focused our analysis on the robustness of the discovered algorithms when tested on a new problem $P^\prime$ and/or with a new budget $B$.

\smallparagraph{Presentation of the results} For each domain, we present the results in a table in which the algorithms have been sorted according to their \textit{testing} scores on $\mathcal{D}_P$. In each column of these tables, we underline both the best generic algorithm and the best discovered algorithm and show in bold all cases in which a discovered algorithm outperforms all tested generic algorithms. We furthermore performed an unpaired t-test between each discovered algorithm and the best generic algorithm. We display significant results  ($p$-value lower than 0.05) by circumscribing them with stars. As in Table \ref{tbl:AlgorithmsUpToDepth3}, we use $sim$ as an abbreviation for $simulate(\pi^{simu})$ in this section.

\subsection{Sudoku}
\label{sec:sudoku}
Sudoku, a Japanese term meaning ``singular number'', is a popular puzzle played around the world. The Sudoku puzzle is made of a grid of $G^2 \times G^2$ cells, which is structured into blocks of size $G \times G$. When starting the puzzle, some cells are already filled in and the objective is to fill in the remaining cells with the numbers $1$ through $G^2$ so that
\begin{itemize}
\item {no row contains two instances of the same number,}
\item {no column contains two instances of the same number,}
\item {no block contains two instances of the same number.}
\end{itemize}

Sudoku is of particular interest in our case because each Sudoku grid corresponds to a different initial state $x_1$. Thus, a good algorithm $A(\cdot)$ is one that intrinsically has the versatility to face a wide variety of Sudoku grids. 

In our implementation, we maintain for each cell the list of numbers that could be put in that cell without violating any of the three previous rules. If one of these lists becomes empty then the grid cannot be solved and we pass to a final state (see Footnote 2). Otherwise, we select the subset of cells whose number-list has the lowest cardinality, and define one action $u \in \mathcal{U}_x$ per possible number in each of these cells (as in \cite{nestedMC2009caze}). The reward associated to a final state is its proportion of filled cells, hence a reward of $1$ is associated to a perfectly filled grid. 

\paragraph{Algorithm discovery} We sample the initial states $x_1$ by filling $33\%$ randomly selected cells as proposed in \cite{nestedMC2009caze}. We denote by Sudoku(G) the distribution over Sudoku problems obtained with this procedure (in the case of $G^2 \times G^2$ games). Even though Sudoku is most usually played with $G=3$ \cite{geem2007harmony}, we carry out the algorithm discovery with $G=4$  to make the problem more difficult. Our training distribution was thus $\mathcal{D}_P= $ Sudoku(4) and we used a training budget of $B=1,000$ evaluations. To evaluate the performance and robustness of the algorithms found, we tested the MCS algorithms on two distributions: $\mathcal{D}_P= $ Sudoku(4) and $\mathcal{D}^\prime_P=$ Sudoku(5), using a budget of $B=1,000$.

\begin{table*}[bt]
\caption{Ranking and Robustness of Algorithms Discovered when Applied to Sudoku}
\begin{center}
\begin{tabular}{l|l|c|c|c}
Name & Search Component & Rank & Sudoku($4$) & Sudoku($5$) \\ \hline
Dis$\#$8 & step(select(repeat(select(sim, 0.5), 5), 0))   & 1  & \textbf{\underline{198.9}} & 487.2 \\
Dis$\#$2 & step(repeat(step(repeat(sim, 5)), 10))         & 2  & \textbf{198.8} & 486.2 \\
Dis$\#$6 & step(\textbf{step(repeat(select(sim, 0), 5))})          & 2  & \textbf{198.8} & 486.2 \\
uct(0)								                         & & 4 & \underline{198.7} & \underline{494.4}  \\
uct(0.3)							                         & & 4 & \underline{198.7} & 493.3 \\
Dis$\#$7 & lookahead(\textbf{step(repeat(select(sim, 0.3), 5))})   & 6 & 198.6 & 486.4 \\
uct(0.5)							                         & & 6 & 198.6 & 492.7 \\
Dis$\#$1 & select(\textbf{step(repeat(select(sim, 1), 5)}), 1)     & 6 & 198.6 & 485.7  \\
Dis$\#$10 & select(\textbf{step(repeat(select(sim, 0.3), 5))})       & 9  & 198.5 & 485.9 \\
Dis$\#$3 & step(select(step(sim), 1))                     & 10 & 198.4 & \underline{493.7} \\
Dis$\#$4 & step(step(step(select(sim, 0.5))))             & 11 & 198.3 & 493.4  \\
Dis$\#$5 & select(step(repeat(sim, 5)), 0.5)              & 11 & 198.3 & 486.3 \\
Dis$\#$9 & lookahead(step(step(select(sim, 1))))          & 13 & 198.1 &  492.8 \\
uct(1)	                                                     & & 13 & 198.1 & 486.9  \\
nmc(3)                                                       & & 15 & 196.7 & 429.7\\
la(1)                                                        & & 16 & 195.6 & 430.1\\
nmc(4)                                                       & & 17 & 195.4 & 430.4 \\
nmc(2)                                                       & & 18 & 195.3 & 430.3\\
nmc(5)                                                       & & 19 & 191.3 & 426.8 \\
la(2)                                                        & & 20 & 174.4 & 391.1 \\
la(4)                                                        & & 21 & 169.2 & 388.5 \\
is                                                           & & 22 & 169.1 & 388.5 \\
la(5)                                                        & & 23 & 168.3 & 386.9 \\
la(3)                                                        & & 24 & 167.1 & 389.1 \\
\multicolumn{5}{c}{|}
\end{tabular}
\end{center}
\label{tb:sudoku}
\end{table*}

Table \ref{tb:sudoku} presents the results, where the scores are the average number of filled cells, which is given by the reward times the total number of cells $G^4$. 
The best generic algorithms on Sudoku(4) are $uct(0)$ and $uct(0.3)$, with an average score of $198.7$. We discover three algorithms that have a better average score ($198.8$ and $198.9$) than $uct(0)$, but, due to a very large variance on this problem (some Sudoku grids are far more easy than others), we could not show this difference to be significant. Although the discovered algorithms are not significantly better than $uct(0)$, none of them is significantly worst than this baseline. Furthermore, all ten discovered algorithms are significantly better than all the other non-uct baselines. Interestingly, four out of the ten discovered algorithms rely on the $uct$ pattern -- $step(repeat(select(sim, \cdot), \cdot))$ -- as shown in bold in the table.

When running the algorithms on the Sudoku(5) games, the best algorithm is still $uct(0)$, with an average score of $494.4$. This score is slightly above the score of the best discovered algorithm ($493.7$). However, all ten discovered algorithms are still significantly better than the non-uct generic algorithms. This shows that good algorithms with Sudoku(4) are still reasonably good for Sudoku(5).

\begin{table}[bt]
\caption{Repeatability Analysis}
\begin{center}
\begin{tabular}{l|c}
Search Component Structure &  Occurrences in the top-ten  \\ \hline
select(\textbf{step(repeat(select(sim)))}) & 11 \\
step(\textbf{step(repeat(select(sim)))}) & 6 \\
step(select(repeat(select(sim)))) & 5 \\
\textbf{step(repeat(select(sim)))} & 5 \\
select(step(repeat(sim))) & 2 \\
select(step(select(repeat(sim))))  & 2 \\
step(select(step(select(sim)))) & 2 \\
step(step(select(repeat(sim)))) & 2 \\
step(repeat(step(repeat(sim)))) & 2 \\
lookahead(\textbf{step(repeat(select(sim)))}) & 2 \\
step(repeat(step(repeat(sim)))) & 2 \\
select(repeat(step(repeat(sim)))) & 1 \\
select(step(repeat(sim))) & 1 \\
lookahead(step(step(select(sim)))) & 1 \\
step(step(step(select(sim)))) & 1 \\
step(step(step(repeat(sim)))) & 1 \\
step(repeat(step(select(sim)))) & 1 \\
step(repeat(step(step(sim)))) & 1 \\
step(select(step(sim))) & 1 \\
step(select(repeat(sim))) & 1 \\
\multicolumn{2}{c}{|}
\end{tabular}
\end{center}
\label{tb:sudoku-repeatability}
\end{table}

\paragraph{Repeatability} In order to evaluate the stability of the results produced by the bandit algorithm, we performed five runs of algorithms discovery with different random seeds and compared the resulting top-tens. What we observe is that our space contains a huge number of MCS algorithms performing nearly equivalently on our distribution of Sudoku problems. In consequence, different runs of the discovery algorithm produce different subsets of these nearly equivalent algorithms. Since we observed that small changes in the constants of $repeat$ and $select$ often have a negligible effect, we grouped the discovered algorithms by structure, i.e. by ignoring the precise values of their constants. Table \ref{tb:sudoku-repeatability} reports the number of occurrences of each search component structure among the five top-tens.  We observe that $uct$ was discovered in five cases out of fifty and that the $uct$ pattern is part of $24$ discovered algorithms.

\begin{table*}[bt]
\caption{Algorithms Discovered when Applied to Sudoku with a CPU time budget}
\begin{center}
\begin{tabular}{l|l|c|c|c}
Name & Search Component & Rank & Rank in Table II & Sudoku($4$)  \\ \hline
Dis$\#$1 & select(step(select(step(sim), 0.3)), 0.3)       & 1 & - & \underline{\textbf{197.2}}  \\
Dis$\#$2 & step(repeat(step(step(sim)), 10))                  & 2 & - & \textbf{196.8} \\
Dis$\#$4 & lookahead(select(step(step(sim)), 0.3), 1)  & 3 & - & \textbf{196.1}   \\
Dis$\#$5 & select(lookahead(step(step(sim)), 1), 0.3)  & 4 & - & \textbf{195.9}  \\
Dis$\#$3 & lookahead(select(step(step(sim)), 0), 1)     & 5 & - & \textbf{195.8}  \\
Dis$\#$6 & step(select(step(repeat(sim, 2)), 0.3))         & 6 & - & \textbf{195.3}  \\
Dis$\#$9 & select(step(step(repeat(sim, 2))), 0)            & 7 & - & \textbf{195.2}  \\
Dis$\#$8 & step(step(repeat(sim, 2)))                              & 8 & - & \textbf{194.8} \\
nmc(2)     &                                                                            & 9 & 18 &  \underline{194.7} \\
nmc(3)      &                                                                           & 10 & 15 & 194.5 \\
Dis$\#$7 & step(step(select(step(sim), 0)))                    & 10 & - & 194.5  \\
Dis$\#$10 & step(repeat(step(step(sim)), 100))            & 10 & - & 194.5  \\
la(1)                                                                             &       & 13 & 16 & 194.2 \\
nmc(4)                                                                         &       & 14 & 17 & 193.7 \\
nmc(5)                                                                         &       & 15 & 19 & 191.4  \\
uct(0.3)							          &        & 16 & 4 & 189.7  \\
uct(0)								 &        & 17 & 4 & 189.4   \\
uct(0.5)							         &         & 18 & 6 & 188.9  \\
uct(1)	                                                &                          & 19 & 13 & 188.8   \\
la(2)                                                        &                          & 20 & 20 & 175.3  \\
la(3)                                                        &                          & 21 & 24 & 170.3  \\
la(4)                                                        &                          & 22 & 21 & 169.3 \\
la(5)                                                        &                          & 23 & 23 & 168.0 \\
is                                                              &                          & 24 & 22 & 167.8 \\
\multicolumn{4}{c}{|}
\end{tabular}
\end{center}
\label{tb:sudoku-time}
\end{table*}

\paragraph{Time-based budget} Since we expressed the budget as the number of calls to the reward function $g(\cdot)$, algorithms that take more time to select their actions may be favored. To evaluate the extent of this potential bias, we performed an experiment by setting the budget to a fixed amount of CPU time. With our C++ implementation, on a 1.9 Ghz computer, about $\approx 350$  Sudoku(4) random simulations can be performed per second. In order to have comparable results with those obtained previously, we thus set our budget to $B=\frac{1000}{350} \approx 2.8$ seconds, during both algorithm discovery and evaluation.

Table \ref{tb:sudoku-time} reports the results we obtain with a budget expressed as a fixed amount of CPU time. For each algorithm, we indicate also its rank in Table \ref{tb:sudoku}. The new best generic algorithm is now $nmc(2)$  and eight out of the ten discovered have a better average score than this generic algorithm. In general, we observe that time-based budget favors $nmc(\cdot)$ algorithms and  decreases the rank of $uct(\cdot)$ algorithms.

In order to better understand the differences between the algorithms found with an evaluations-based budget and those found with a time-based budget, we counted the number of occurrences of each of the search components among the ten discovered algorithms in both cases. These counts are reported in Table \ref{tb:sudoku-composition}. We observe that the time-based budget favors the $step$ search component, while reducing the use of $select$. This can be explained by the fact that $select$ is our search component that involves the most extra-computational cost, related to the storage and the manipulation of the game tree.

\subsection{Real Valued Symbolic Regression}
\label{sec:symbreg}
Symbolic Regression consists in searching in a large space of symbolic expressions for the one that best fits a given regression dataset. Usually this problem is treated using Genetic Programming approaches. In the line of \cite{cazenave2010nested}, we here consider MCS techniques as an interesting alternative to Genetic Programming. In order to apply MCS techniques, we encode the expressions as sequences of symbols. We adopt the Reverse Polish Notation (RPN) to avoid the use of parentheses. As an example, the sequence $[a,b,+,c,*]$ encodes the expression $(a+b)*c$.
The alphabet of symbols we used is $\{x,1,+,-,*,/,\sin,\cos, \log, \exp , stop \}$. 
The initial state $x_1$ is the empty RPN sequence. Each action $u$ then adds one of these symbols to the sequence. When computing the set of valid actions $\mathcal{U}_x$, we reject symbols that lead to invalid RPN sequences, such as $[+, +, +]$. A final state is reached either when the sequence length is equal to a predefined maximum $T$ or when the symbol $stop$ is played. In our experiments, we performed the training with a maximal length of $T=11$. The reward associated to a final state is equal to $1 - mae$, where $mae$ is the mean absolute error associated to the expression built.

\begin{table}[bt]
\caption{Search Components Composition  Analysis}
\begin{center}
\begin{tabular}{l|c|c}
Name & Evaluations-based Budget & Time-based Budget  \\ \hline
$repeat$ & 8 &  5 \\
$simulate$ & 10 & 10 \\
$select$ & 12 & 8 \\
$step$  & 16 & 23 \\
$lookahead$ & 2 & 3 \\
\multicolumn{3}{c}{|}
\end{tabular}
\end{center}
\label{tb:sudoku-composition}
\end{table}

\begin{table}[tb]
\caption{\label{tab:sr}Symbolic Regression Testbed: target expressions and domains.}
\begin{center}
\begin{tabular}{l|c}
Target Expression $f^P(\cdot)$ &  Domain  \\ \hline
$x^3+x^2+x$ & $[-1,1]$\\%
$x^4+x^3+x^2+x$ & $[-1,1]$\\%
$x^5+x^4+x^3+x^2+x$ & $[-1,1]$\\%
$x^6+x^5+x^4+x^3+x^2+x$ & $[-1,1]$\\%
$\sin(x^2)\cos(x)-1$ &  $[-1,1]$ \\
$\sin(x)+\sin(x+x^2)$ &  $[-1,1]$ \\
$\log(x+1)+\log(x^2+1)$ &  $ [0,2]$ \\
$\sqrt{x}$ &  $ [0,4]$ \\
\multicolumn{2}{c}{|}
\end{tabular}
\end{center}

\end{table}

\begin{table}[tb]
\caption{\label{tab:sr2}Symbolic Regression Robustness Testbed: target expressions and domains.}
\begin{center}
\begin{tabular}{l|c}
Target Expression $f^P(\cdot)$ &  Domain  \\ \hline
$x^3-x^2-x$ & $[-1,1]$\\%
$x^4-x^3-x^2-x$ & $[-1,1]$\\%
$x^4+\sin(x)$ & $[-1,1]$\\%
$\cos(x^3)+\sin(x+1)$ & $[-1,1]$\\%
$\sqrt(x)+x^2$ &  $[0,4]$ \\
$x^6+1$ &  $[-1,1]$ \\
$\sin(x^3+x^2)$ &  $ [-1,1]$ \\
$\log(x^3+1)+x$ &  $ [0,2]$ \\
\multicolumn{2}{c}{|}
\end{tabular}
\end{center}

\end{table}

\begin{table*}[bt]
\caption{Ranking and Robustness of the Algorithms Discovered when Applied to Symbolic Regression}
\begin{center}
\begin{tabular}{l|l|c|c|c|c|c}
Name & Search Component & Rank & $T=11$ & $T=21$ & $T=11$,  $B = 10^5$ & $\mathcal{D}^\prime_P$ \\ \hline
Dis$\#$1 & step(step(lookahead(lookahead(sim))))               & 1 & *\textbf{\underline{0.066}}* & *\textbf{\underline{0.083}}* & *\textbf{\underline{0.036}}* & 0.101 \\
Dis$\#$5 & step(repeat(lookahead(lookahead(sim)), 2))          & 2  & *\textbf{0.069}* & *\textbf{0.085}* &  *\textbf{0.037}* & 0.106 \\
Dis$\#$2 & step(lookahead(lookahead(repeat(sim, 2))))          & 2 & *\textbf{0.069}* & *\textbf{0.084}* & *\textbf{0.038}* & \textbf{0.100} \\
Dis$\#$8 & step(lookahead(repeat(lookahead(sim), 2)))          & 2  & *\textbf{0.069}* & *\textbf{0.084}* &  *\textbf{0.040}* & 0.112 \\
Dis$\#$7 & step(lookahead(lookahead(select(sim, 1))))          & 5  & *\textbf{0.070}* & 0.087 &  *\textbf{0.040}* & 0.103 \\
Dis$\#$6 & step(lookahead(lookahead(select(sim, 0))))          & 6  & \textbf{0.071} & 0.087 &  *\textbf{0.039}* & 0.110 \\
Dis$\#$4 & step(lookahead(select(lookahead(sim), 0)))          & 6  & \textbf{0.071} & 0.087 &  *\textbf{0.038}* & 0.101 \\
Dis$\#$3 & step(lookahead(lookahead(sim)))                     & 6  & \textbf{0.071} & \textbf{0.086} & 0.056 & \textbf{0.100} \\
la(2)                                                             & & 6 & \underline{0.071} & \underline{0.086} & 0.056 & \underline{0.100}  \\
Dis$\#$10 & step(lookahead(select(lookahead(sim), 0.3)))       & 10  & 0.072 & 0.088 &  *\textbf{0.040}* & 0.108 \\
la(3)                                                             & & 11 & 0.073 & 0.090 & \underline{0.053}  & 0.101 \\
Dis$\#$9 & step(repeat(select(lookahead(sim), 0.3), 5))        & 12  & 0.077 & 0.091 &  *\textbf{0.048}* & *\underline{\textbf{0.099}}* \\
nmc(2)                                                            & & 13 & 0.081 & 0.103 & 0.054 & 0.109  \\
nmc(3)                                                            & & 14 & 0.084 & 0.104 & \underline{0.053} & 0.118 \\
la(4)                                                             & & 15 & 0.088 & 0.116 & 0.057 & 0.101 \\
nmc(4)                                                            & & 16 & 0.094 & 0.108 & 0.059 & 0.141  \\
la(1)                                                             & & 17 & 0.098 & 0.116 & 0.066 & 0.119 \\
la(5)                                                             & & 18 & 0.099 & 0.124 & 0.058 & 0.101 \\
is                                                                & & 19 & 0.119 & 0.144 & 0.087 & 0.139 \\
nmc(5)                                                            & & 20 & 0.120 & 0.124 & 0.069 & 0.140 \\
uct(0)								                              & & 21 & 0.159 & 0.135 & 0.124 & 0.185 \\
uct(1)								                              & & 22 & 0.147 & 0.118 & 0.118& 0.161 \\
uct(0.3)						                                  & & 23 & 0.156 & 0.112 & 0.135& 0.177 \\
uct(0.5)							                              & & 24 & 0.153 & 0.111 & 0.124 & 0.184 \\

\multicolumn{6}{c}{|}
\end{tabular}
\end{center}
\label{tb:symregResult}
\end{table*}

We used a synthetic benchmark, which is classical in the field of Genetic Programming \cite{uy2011semantically}. To each problem $P$ of this benchmark is associated a target expression $f^P(\cdot) \in \mathbbm{R}$, and the aim is to re-discover this target expression given a finite set of samples $(x, f^P(x))$. Table \ref{tab:sr} illustrates these target expressions. In each case, we used 20 samples $(x, f^P(x))$, where $x$ was obtained by taking uniformly spaced elements from the indicated domains. The training distribution $\mathcal{D}_P$ was the uniform distribution over the eight problems given in Table \ref{tab:sr}.

The training budget was $B = 10,000$. We evaluate the robustness of the algorithms found in three different ways: by changing the maximal length $T$ from 11 to 21, by increasing the budget $B$ from 10,000 to 100,000 and by testing them on another distribution of problems $\mathcal{D}^\prime_P$. The distribution $\mathcal{D}^\prime_P$ is the uniform distribution over the eight new problems given in Table \ref{tab:sr2}.

The results are shown in Table \ref{tb:symregResult}, where we report directly the $mae$ scores (lower is better). The best generic algorithm is $la(2)$ and corresponds to one of the discovered algorithms (Dis$\#$3). Five of the discovered algorithms significantly outperform this baseline with scores down to $0.066$. Except one of them, all discovered algorithms rely on two nested $lookahead$ components and generalize in some way the $la(2)$ algorithm.

When setting the maximal length to $T=21$, the best generic algorithm is again $la(2)$ and we have four discovered algorithms that still significantly outperform it. When increasing the testing budget to $B=100,000$, nine discovered algorithms out of the ten significantly outperform the best generic algorithms, $la(3)$ and $nmc(3)$. These results thus show that the algorithms discovered by our approach are robust both w.r.t. the maximal length $T$ and the budget $B$.

In our last experiment with the distribution $\mathcal{D}^\prime_P$, there is a single discovered algorithm that significantly outperform $la(2)$. However, all ten algorithms behave still reasonably well and significantly better than the non-lookahead generic algorithms. This result is particularly interesting since it shows that our approach was able to discover algorithms that work well for symbolic regression in general, not only for some particular problems.

\subsection{Morpion Solitaire}
\label{sec:morpion}

\begin{figure*}[tb]
		\includegraphics[scale=0.24, clip]{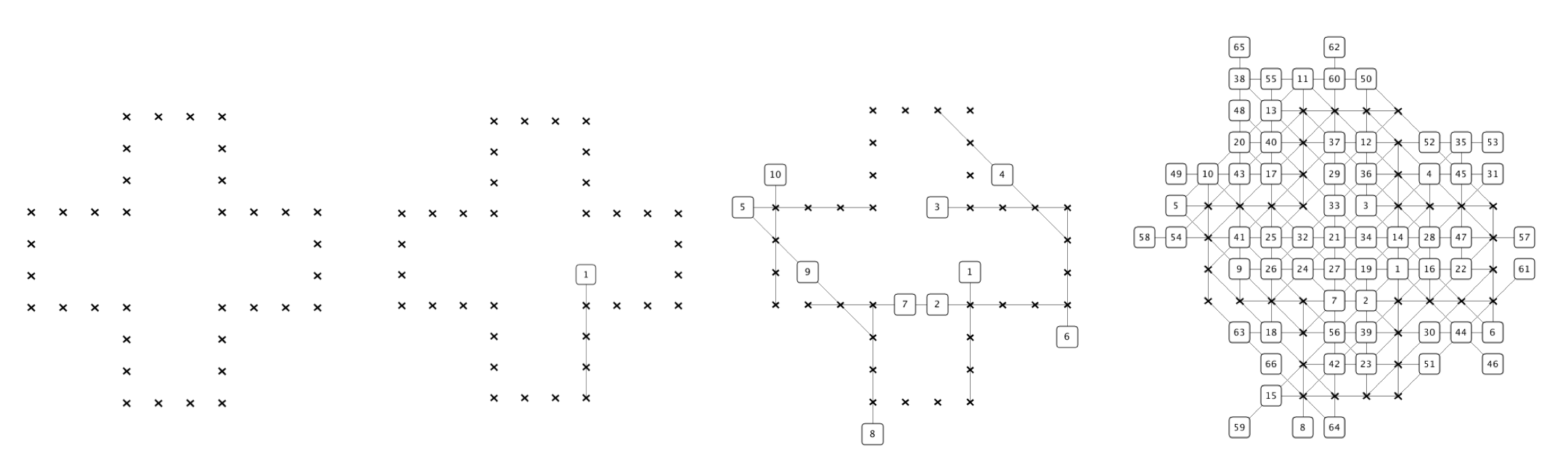}
	\caption{A random policy that plays the game Morpion Solitaire 5T: initial grid; after 1 move; after 10 moves; game end.}
	\label{fig:morpion}
\end{figure*}

The classic game of morpion solitaire \cite{morpion} is a single player, pencil and paper game, whose world record has been improved several times over the past few years using MCS techniques \cite{rosin2011nested,nestedMC2009caze,cazenave2007reflexive}. 
This game is illustrated in Figure \ref{fig:morpion}. The initial state $x_1$ is an empty cross of points drawn on the intersections of the grid. Each action places a new point at a grid intersection in such a way that it forms a new line segment connecting consecutive points that include the new one.  New lines can be drawn horizontally, vertically, and diagonally. The game is over when no further actions can be taken. The goal of the game is to maximize the number of lines drawn before the game ends, hence the reward associated to final states is this number\footnote{In practice, we normalize this reward by dividing it by $100$ to make it approximately fit into the range $[0,1]$. Thanks to this normalization, we can keep using the same constants for both the UCB policy used in the algorithm discovery and the UCB policy used in $select$.}.

There exist two variants of the game: ``Disjoint'' and ``Touching''. ``Touching'' allows parallel lines to share an endpoint, whereas ``Disjoint'' does not. Line segments with different directions are always permitted to share points. The game is NP-hard \cite{demaine2006morpion} and presumed to be infinite under certain configurations. In this paper, we treat the $5D$ and $5T$ versions of the game, where $5$ is the number of consecutive points to form a line, $D$ means disjoint, and $T$ means touching. 

We performed the algorithm discovery in a ``single training problem'' scenario: the training distribution $\mathcal{D}_P$ always returns the same problem $P$, corresponding to the $5T$ version of the game. The initial state of $P$ was the one given in the leftmost part of Figure \ref{fig:morpion}. The training budget was set to $B=10,000$. To evaluate the robustness of the algorithms, we, on the one hand, evaluated them on the $5D$ variant of the problem and, on the other hand, changed the evaluation budget from 10,000 to 100,000. The former provides a partial answer to how rule-dependent these algorithms are, while the latter gives insight into the impact of the budget on the algorithms' ranking. 

\begin{table*}[bt]
\caption{Ranking and Robustness of Algorithms Discovered when Applied to Morpion}
\begin{center}
\begin{tabular}{l|l|c|c|c|c}
Name & Search Component & Rank & $5T$ & $5D$ & $5T, B = 10^5$ \\ \hline
Dis$\#$1 & step(select(step(simulate),0.5))                        & 1  & *\textbf{\underline{91.24}}* & *\textbf{63.66}* & *\textbf{\underline{97.28}}* \\
Dis$\#$4 & step(select(step(select(sim,0.5)),0))       & 2  & *\textbf{91.23}* & *\textbf{63.64}* & *\textbf{96.12}* \\
Dis$\#$3 & step(select(step(select(sim,1.0)),0))       & 3  & *\textbf{91.22}* & *\textbf{63.63}* & *\textbf{96.02}* \\
Dis$\#$2 & step(step(select(sim,0)))                   & 4  & *\textbf{91.18}* & *\textbf{63.63}* & *\textbf{96.78}* \\
Dis$\#$8 & step(select(step(step(sim)),1))             & 5  & *\textbf{91.12}* & *\textbf{63.63}* & *\textbf{96.67}* \\
Dis$\#$9 & step(select(step(select(sim,0)),0.3))       & 6  & *\textbf{91.22}* & *\textbf{63.67}* & *\textbf{96.02}* \\
Dis$\#$5 & select(step(select(step(sim),1.0)),0)       & 7  & *\textbf{91.16}* & *\textbf{63.65}* & *\textbf{95.79}* \\
Dis$\#$10 & step(select(step(select(sim,1.0)),0.0))    & 8  & *\textbf{91.21}* & *\textbf{63.62}* & *\textbf{95.99}* \\
Dis$\#$6 & lookahead(step(step(sim)))                  & 9 & *\textbf{91.15}* & *\textbf{\underline{63.68}}* & *\textbf{96.41}* \\
Dis$\#$7 & lookahead(step(step(select(sim, 0))))       & 10 & *\textbf{91.08}* & \textbf{63.67} & *\textbf{96.31}* \\
la(1)                                                     & & 11 & \underline{90.63} & 63.41 & 95.09\\
nmc(3)                                                    & & 12 & 90.61 & 63.44 & \underline{95.59}\\
nmc(2)                                                    & & 13 & 90.58 & \underline{63.47} & 94.98\\
nmc(4)                                                    & & 14 & 90.57 & 63.43 & 95.24\\
nmc(5)                                                    & & 15 & 90.53 & 63.42 & 95.17\\
uct(0)						& & 16 & 89.40 & 63.02 & 92.65 \\
uct(0.5)				     & & 17 & 89.19 & 62.91 & 92.21 \\
uct(1)					      & & 18 & 89.11 & 63.12 & 92.83 \\
uct(0.3)				     & & 19 & 88.99 & 63.03 & 92.32 \\
la(2)                                                     & & 20 & 85.99 & 62.67 & 94.54\\
la(3)                                                     & & 21 & 85.29 & 61.52 & 89.56\\
is                                                        & & 21 & 85.28 & 61.40 &  88.83\\
la(4)                                                     & & 23 & 85.27 & 61.53 & 88.12\\
mcts                                                      & & 24 & 85.26 & 61.48 & 89.46\\
la(5)                                                     & & 25 & 85.12 & 61.52 & 87.69\\
\multicolumn{6}{c}{|}
\end{tabular}
\end{center}
\label{tb:morpion}
\end{table*}

The results of our experiments on Morpion Solitaire are given in Table \ref{tb:morpion}. Our approach proves to be particularly successful on this domain: each of the ten discovered algorithms significantly outperforms all tested generic algorithm. Among the generic algorithms, $la(1)$ gave the best results (90.63), which is 0.46 below the worst of the ten discovered algorithms.

When moving to the $5D$ rules, we observe that all ten discovered algorithms still significantly outperform the best generic algorithm. This is particularly impressive, since it is known that the structure of good solutions strongly differs between the $5D$ and $5T$ versions of the game  \cite{morpion}. The last column of Table \ref{tb:morpion} gives the performance of the algorithms with budget $B=10^5$. We observe that all ten discovered algorithms also significantly outperform the best generic algorithm in this case. Furthermore, the increase in the budget seems to also increase the gap between the discovered and the generic algorithms.


\subsection{Discussion}
\label{sec:xp:dis}

We have seen that on each of our three testbeds, we discovered algorithms, which are competitive with, or even significantly better than generic ones. This demonstrates that our approach is able to generate new MCS algorithms specifically tailored to the given class of problems. We have performed a study of the robustness of these algorithms by either changing the problem  distribution or by varying the budget $B$, and found that the algorithms discovered can outperform generic algorithms even on problems significantly different from those used for the training. 

The importance of each component of the grammar depends heavily on the problem. For instance, in Symbolic Regression, all ten best algorithms discovered rely on two nested $lookahead$ components, whereas in Sudoku and Morpion, $step$ and $select$ appear in the majority of the best algorithms discovered.

\section{Related Work}
\label{sec:related}
Methods for automatically discovering MCS algorithms can be characterized through three main components: the space of candidate algorithms, the performance criterion, and the search method for finding the best element in the space of candidate algorithms. 

Usually, researchers consider spaces of candidate algorithms that only differ in the values of their constants. In such a context, the problem amounts to tuning the constants of a generic MCS algorithm. Most of the research related to the tuning of these constants takes as performance criterion the mean score of the algorithm over the distribution of target problems. Many search algorithms have been proposed for computing the best constants. For instance, \cite{perrick2012tron} employs a grid search approach combined with self-playing, \cite{chaslot2008parameter} uses cross-entropy as a search method to tune an agent playing GO,  \cite{coulomClop} presents a generic black-box optimization method based on local quadratic regression, \cite{Maes2011EwrlLookahead} uses Estimation Distribution Algorithms with Gaussian distributions, \cite{chapelle2011empirical} uses Thompson Sampling, and \cite{bourki2010parameter} uses, as in the present paper, a multi-armed bandit approach. The paper \cite{berthier2010consistency} studies the influence of the tuning of MCS algorithms on their asymptotic consistency and shows that pathological behaviour may occur with tuning. It also proposes a tuning method to avoid such behaviour.

Research papers that have reported empirical evaluations of several MCS algorithms in order to find the best one are also related to this automatic discovery problem. The space of candidate algorithms in such cases is the set of algorithms they compare, and the search method is an exhaustive search procedure. As a few examples, \cite{perrick2012tron} reports on a comparison between algorithms that differ in their selection policy, \cite{gelly2007combining} and \cite{chaslot2008progressive} compare improvements of the UCT algorithm (RAVE and progressive bias) with the original one on the game of GO, and \cite{onexp32011} evaluates different versions of a two-player MCS algorithm on generic sparse bandit problems. \cite{brownesurvey} provides an in-depth review of different MCS algorithms and their successes in different applications.

The main feature of the approach proposed in the present paper is that it builds the space of candidate algorithms by using a rich grammar over the search components. In this sense, \cite{cazenave2007evolving,maes2011automatic} are certainly the papers  which are the closest to ours, since they also use a grammar to define a search space, for, respectively, two player games and multi-armed bandit problems. However, in both cases, this grammar only models a selection policy and is made of classic functions such as $+$, $-$, $*$, $/$, $\log{}$, $\exp{}$, and $\sqrt{}$. We have taken one step forward, by directly defining a grammar over the MCS algorithms that covers very different MCS techniques. Note that the search technique of \cite{cazenave2007evolving} is based on genetic programming. 

The decision as to what to use as the performance criterion is not as trivial as it looks, especially for multi-player games, where opponent modelling is crucial for improving over game-theoretically optimal play \cite{billings2002challenge}. For example, the maximization of the victory rate or loss minimization against a wide variety of opponents for a specific game can lead to different choices of algorithms. Other examples of criteria to discriminate between algorithms are simple regret \cite{bourki2010parameter} and the expected performance over a distribution density \cite{nannen2007relevance}.

\section{Conclusion}
\label{sec:conclusion}
In this paper we have addressed the problem of automatically identifying new Monte Carlo search (MCS) algorithms performing well on a distribution of training problems. To do so, we introduced a grammar over the MCS algorithms that generates a rich space of candidate algorithms (and which describes, along the way, using a particularly compact and elegant description, several well-known MCS algorithms). To efficiently search inside this space of candidate algorithms for the one(s) having the best average performance on the training problems, we relied on a multi-armed bandit type of optimisation algorithm.

Our approach was tested on three different domains: Sudoku, Morpion Solitaire, and Symbolic Regression. The results showed that the algorithms discovered this way often significantly outperform generic algorithms such as UCT or NMC. Moreover, we showed that they had good robustness properties, by changing the testing budget and/or by using a testing problem distribution different from the training distribution.

This work can be extended in several ways. For the time being, we used the mean performance over a set of training problems to discriminate between different candidate algorithms.  One direction for future work would be to adapt our general approach to use other criteria, e.g., worst case performance measures. In its current form, our grammar only allows using predefined simulation policies. Since the simulation policy typically has a major impact on the performance of a MCS algorithm, it could be interesting to extend our grammar so that it could also ``generate'' new simulation policies. This could be arranged by adding a set of simulation policy generators in the spirit of our current search component generators. Previous work has also demonstrated that the choice of the selection policy could have a major impact on the performance of Monte Carlo tree search algorithms. Automatically generating selection policies is thus also a direction for future work. Of course, working with richer grammars will lead to larger candidate algorithm spaces, which in turn, may require developing more efficient search methods than the multi-armed bandit one used in this paper.  Finally, another important direction for future research is to extend our approach to more general settings than single-player games with full observability.

\section*{Acknowledgment}
This paper presents research results of the Belgian Network DYSCO (Dynamical Systems, Control, and Optimization), funded by the Interuniversity Attraction Poles Programme, initiated by the Belgian State, Science Policy Office. The scientific responsibility rests with its author(s).

\ifCLASSOPTIONcaptionsoff
  \newpage
\fi

\bibliographystyle{IEEEtran}

\bibliography{IEEEabr,ciaig}

\end{document}